\documentclass[runningheads]{llncs}

\usepackage{eccv}
\usepackage{verbatim}

\usepackage{algorithm} %
\usepackage{algorithmic} %
\usepackage{multirow}

\usepackage{eccvabbrv}
\usepackage{xcolor}
\definecolor{shapecolor}{rgb}{0.0,0.5,0.0}

\definecolor{dkgreen}{rgb}{0,0.6,0}
\definecolor{gray}{rgb}{0.5,0.5,0.5}
\definecolor{mauve}{rgb}{0.58,0,0.82}

\usepackage{listings}

\usepackage{graphicx}
\usepackage{booktabs}
\usepackage{multicol}
\usepackage[accsupp]{axessibility} 

\usepackage[pagebackref,breaklinks,colorlinks,citecolor=eccvblue]{hyperref}
\usepackage{orcidlink}

\begin{document}

\title{Pan-Mamba: Effective pan-sharpening with State Space Model
} 

\titlerunning{Pan-Mamba: Effective pan-sharpening with State Space }

\author{Xuanhua He\inst{1,2} \and
 Ke Cao\inst{1,2} \and
Keyu Yan \inst{1,2} \and Rui Li \inst{1} \and Chengjun Xie \inst{1} \and Jie Zhang \inst{1} \and Man Zhou \inst{2} }

\authorrunning{X.~He et al.}

\institute{Hefei Institutes of Physical Science, Chinese Academy of Sciences \and
University of Science and Technology of China}

\maketitle

\begin{abstract}
Pan-sharpening involves integrating information from low-resolution multi-spectral and high-resolution panchromatic images to generate high-resolution multi-spectral counterparts. While recent advancements in the state space model, particularly the efficient long-range dependency modeling achieved by Mamba, have revolutionized computer vision community, its untapped potential in pan-sharpening motivates our exploration. Our contribution, Pan-Mamba, represents a novel pan-sharpening network that leverages the efficiency of the Mamba model in global information modeling. In Pan-Mamba, we customize two core components: channel swapping Mamba and cross-modal Mamba, strategically designed for efficient cross-modal information exchange and fusion. The former initiates a lightweight cross-modal interaction through the exchange of partial panchromatic and multi-spectral channels, while the latter facilities the information representation capability by exploiting inherent cross-modal relationships. Through extensive experiments across diverse datasets, our proposed approach surpasses state-of-the-art methods, showcasing superior fusion results in pan-sharpening. To the best of our knowledge, this work is the first attempt in exploring the potential of the Mamba model and establishes a new frontier in the pan-sharpening techniques. The source code is available at \url{https://github.com/alexhe101/Pan-Mamba}.

  \keywords{Pan-sharpening \and Mamba}
\end{abstract}

\section{Introduction}
\label{sec:intro}

With the rapid development of satellite technology, there exists a great need for high-resolution multi-spectral (HRMS) remote sensing image across various fields. However, owing to constraints imposed by physical sensors, acquiring high-resolution multi-spectral images directly through remote sensing satellites poses a considerable challenge. Therefore, satellites are equipped with two distinct types of panchromatic (PAN) and multi-spectral sensors, designed to capture low-resolution multi-spectral (LRMS) images and high-resolution texture-rich PAN images. These two sets of images with complementary information are subsequently fused by pan-sharpening technique, which enables the acquisition of high-resolution multi-spectral image.

In recent years, pan-sharpening technology has garnered considerable attention. Classical pan-sharpening approaches  heavily leaned on manually designed physical rules~\cite{SFIM,Brovey} and encountered difficulties arising from insufficient representation, ultimately leading to suboptimal results. Inspired by the powerful learning capability of deep learning, the pioneer PNN~\cite{pnn} marked a pivotal moment by integration of deep learning methods into this domain, showcasing great advancements over traditional techniques. Subsequently, a flood of increasingly complex deep learning-based models has been introduced, encompassing multi-scale methods~\cite{msdcnn}, models leveraging frequency domain information~\cite{zhou2022spatial,he2023multi}, architectures based on Transformer models~\cite{zhou2022panformer}, models incorporating flow model~\cite{yang2023panflownet}, and those driven by  domain-specific  prior knowledge~\cite{gppnn}.

Nevertheless, aforementioned promising methods exhibit certain limitations that hinder further performance improvement. Specifically, the first challenge lies in capturing the global information. INNformer~\cite{zhou2022pan} and Panformer~\cite{zhou2022panformer} attempt to model global information by incorporating vision transformer~\cite{vit} and Swin Transformer blocks~\cite{liu2021swin}, respectively. However, the former introduces computational complexity, rendering its application challenging, while the latter's window partitioning imposes constraints on the model's receptive field and disrupts the locality of features~\cite{xiao2023random}. SFINet~\cite{zhou2022spatial} and MSDDN~\cite{he2023multi} adopt a different approach by introducing Fourier transform to model global information. However, the interaction between the frequency domain and spatial domain introduces information gaps~\cite{yu2024deep}, and the fixed convolution parameters hinder the model's adaptive ability to varying inputs. Conversely, some techniques that aim to reduce the complexity of self-attention, such as window partitioning~\cite{liu2021swin} and transposed self-attention~\cite{zamir2022restormer}, sacrifice to some extent the inherent capabilities of self-attention, including global information modeling and input adaptability. Secondly, current methods exhibit shortcomings in adequate and efficient information fusion. Panformer employs window-based cross attention, INNformer and PanFlowNet rely on invertible neural networks, and SFINET utilizes a Fourier convolution for feature fusion. However, these approaches fall short in fully capturing the cross-modal correlations between the two modalities and suffer from high computational complexity.
The birth of the Mamba~\cite{gu2023mamba} offers a novel solution to the aforementioned challenges. It features input-adaptive and global information modeling capabilities akin to self-attention, while maintaining linear complexity, reduced computational overhead, and enhanced inference speed. Notably, in the realm of natural language processing, the Mamba model has demonstrated superior results compared to the Transformer architecture.

Considering these considerations, our approach focuses on enhancing models through two key perspectives: feature extraction and feature fusion. We introduce Pan-Mamba, a pan-sharpening network that leverages Mamba as the core module.  Our design includes channel swapping Mamba and cross modal Mamba for global information extraction and efficient feature fusion from both PAN and LRMS images. The channel swapping Mamba initiates a preliminary cross-modal interaction by exchanging partial PAN channels and LRMS channels, facilitating a lightweight and efficient fusion of information. Meanwhile, the cross modal Mamba utilizes the inherent cross-modal relationship between the two, enabling fusion, filtering of redundant modal features, and obtaining refined fusion results. Owing to its efficient feature extraction and fusion capabilities, our model has surpassed state-of-the-art methods, achieving superior fusion results.
Our contributions can be summarized as follows:
\begin{itemize}
    \item This work is the first attempt to introduce the Mamba model into the pan-sharpening domain and presents a novel pan-sharpening network. This approach facilitates efficient long-range information modeling and cross-modal information interaction.
    \item We tailor the channel-swapping Mamba block and cross-modal Mamba block to enable efficient exchange and fusion of cross-modal information.
    \item Conducting extensive experiments across multiple datasets, our proposed method demonstrates state-of-the-art results in both qualitative and quantitative assessments.
\end{itemize}

\section{Related Work}
\subsection{Pan-sharpening}
The methods of pan-sharpening is primarily categorized into two parts: traditional approaches and deep learning-based methods. Traditional methods predominantly rely on manually designed priors, encompassing component substitution algorithms~\cite{Brovey,SFIM,GFPCA,IHS}, multi-resolution analysis algorithms~\cite{HPF,ATWT1999}, and variational optimization algorithms~\cite{fasbender2008bayesian}. Component substitution algorithms leverage the spatial details of PAN images to replace the spatial information of LRMS images. Multi-resolution methods conduct multi-resolution analysis and subsequently fuse the two images, whereas variational optimization-based algorithms model the fusion process as an energy function and iteratively solve it. These methods suffer from limitations in performance due to their insufficient feature representation.

The rise of deep learning in pan-sharpening was initiated by the PNN~\cite{pnn} model, which, drawing inspiration from SRCNN~\cite{srcnn}, devised a simple three-layer neural network with promising outcomes. Subsequent advancements introduced more complex designs to this domain, such as PanNet~\cite{yang2017pannet} utilizing ResNet blocks to capture high-frequency information, MSDCNN~\cite{msdcnn} introducing multi-scale convolution for processing the multi-scale structure of remote sensing images, and SRPPNN~\cite{srppnn} employing a progressive upsampling strategy. The emergence of Transformers has influenced pan-sharpening, with INNformers~\cite{zhou2022pan} and Panfomers~\cite{zhou2022panformer} introducing self-attention mechanisms. SFINet~\cite{zhou2022spatial} and MSDDN~\cite{he2023multi} employ Fourier transforms to capture global features and facilitate the learning of high-frequency information. PanFlowNet~\cite{yang2023panflownet} incorporates flow-based model for processing remote sensing images. Furthermore, prior information-driven methods have been integrated into this field, exemplified by MutNet~\cite{zhou2022mutual} and GPPNN~\cite{gppnn}, which leverage prior knowledge between modalities to promote image fusion.

\subsection{State Space Model}
The concept of the State Space model was initially introduced in the S4~\cite{s4} model, presenting a distinctive architecture capable of effectively modeling global information in comparison to conventional CNN or Transformer architectures. Building upon S4, the S5~\cite{smith2022simplified} model emerged, strategically reducing complexity to a linear level. The subsequent H3~\cite{h3} model further refined and expanded upon this foundation, enabling the model to perform competitively with Transformers in language model tasks. Mamba~\cite{gu2023mamba}, in turn, introduced an input-adaptive mechanism to enhance the State Space model, resulting in higher inference speed, throughput, and overall metrics compared to Transformers of equivalent scale.

The application of the State Space model extended into visual tasks with the introduction of Vision Mamba~\cite{zhu2024vision} and Vmaba~\cite{liu2024vmamba}. These adaptations yielded commendable results in classification and segmentation tasks, successfully penetrating fields such as medical image segmentation~\cite{xing2024segmamba,ma2024u}. Notably, the potential of this model in multimodal image fusion remains an area that has not been thoroughly explored.

\section{Methods}
In this section, we begin by introducing the fundamental knowledge of the state space model. Subsequently, we delve into a detailed exploration of our model, encompassing its architectural framework, module design, and our loss function.

\subsection{Preliminaries}
The state space sequence model and Mamba draw inspiration from linear systems, with the goal of mapping a one-dimensional function or sequence, denoted as $x(t)\in \mathbf{R}$, to $y(t)$ through the hidden space $h(t)\in \mathbf{R}^{N}$. In this context, $\mathbf{A}\in \mathbf{R}^{N\times N}$ serves as the evolution parameter, while $\mathbf{B}\in \mathbf{R}^{N \times 1}$ and $\mathbf{C}\in\mathbf{R}^{1\times N}$ act as the projection parameters. The system can be mathematically expressed using the following formula.
\begin{align}
    &h'(t) = \mathbf{A}h(t)+\mathbf{B}x(t),\\
    &y(t) = \mathbf{C}h'(t).
\end{align}

The S4 and Mamba models serve as discrete counterparts to the continuous system, incorporating a timescale parameter $\mathbf{\Delta}$ to convert continuous parameters $\mathbf{A}$ and $\mathbf{B}$ into their discrete counterparts $\Bar{\mathbf{A}}$ and $\Bar{\mathbf{B}}$. The prevalent approach employed for this transformation is the zero-order hold (ZOH) method, which can be formally defined as follows:
\begin{align}
    &\Bar{\mathbf{A}} = \mathbf{exp(\Delta A)},\\
    &\Bar{\mathbf{B}} =\mathbf{(\Delta A)}^{-1}\mathbf{(exp(\Delta A)-I)\cdot \Delta B}. \label{e3}
\end{align}
The discrete representation of this linear system can be formulated as follows:
\begin{align}
    &h_t = \Bar{\mathbf{A}}h_{t-1}+\Bar{\mathbf{B}}x_t,\\
    &y_t = \mathbf{C}h_t.
\end{align}
Finally, the output is derived through global convolution:
\begin{align}
    &\Bar{\mathbf{K}} = (\mathbf{C\Bar{B},C\Bar{A}\Bar{B},...,C\Bar{A}^{M-1}\Bar{B}}),\\  \label{e7}
    &\mathbf{y} = x*\mathbf{\Bar{K}}. 
\end{align}
Here, $\mathbf{M}$ denotes the sequence length of x, and $\mathbf{\Bar{K}}\in \mathbf{R}^{M}$ represents a structured convolutional kernel.

\subsection{Network Architecture}
\begin{figure}
    \centering
    \includegraphics[width=\textwidth]{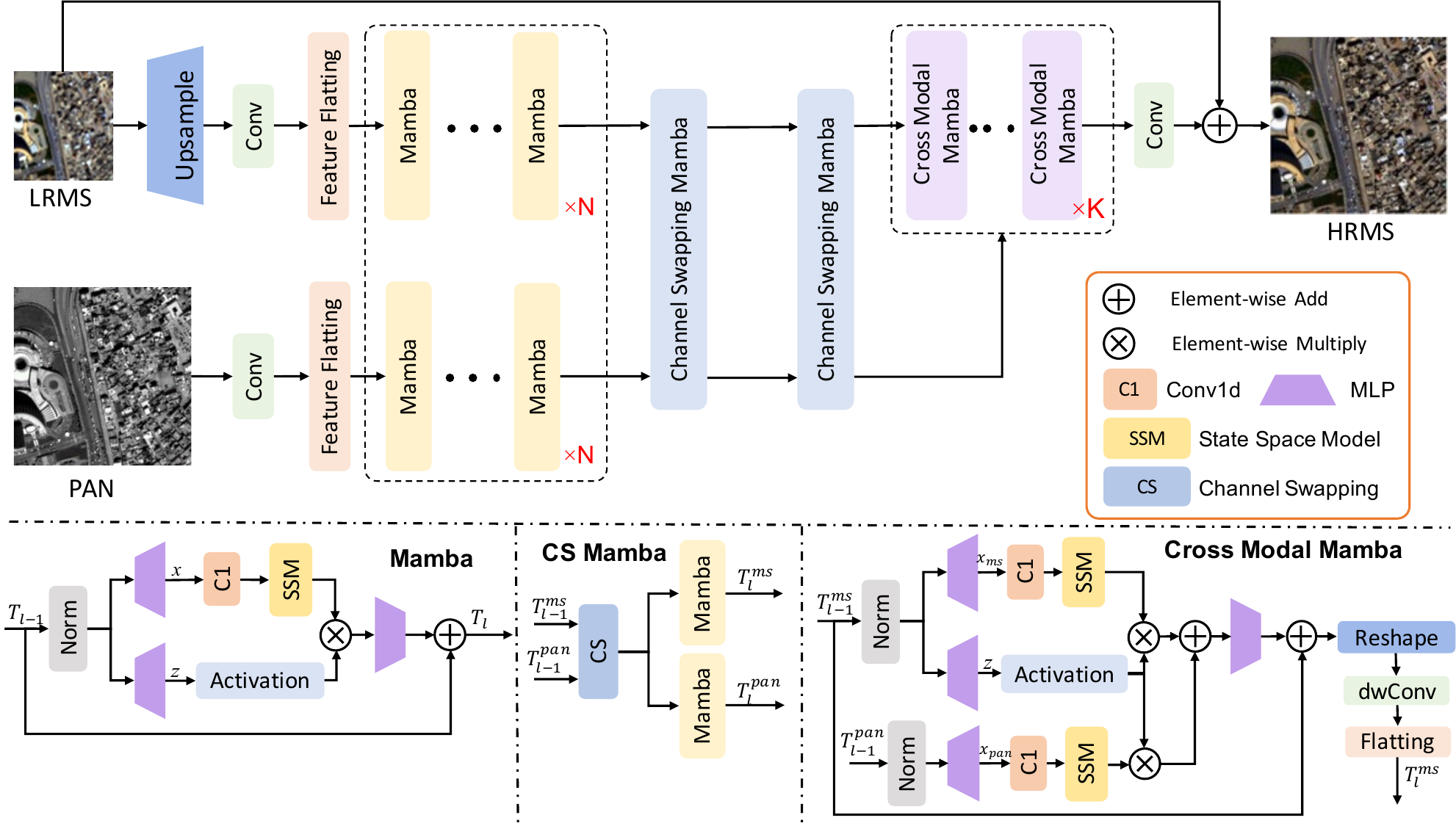}
    \caption{The network structure of our proposed Pan-Mamba, which includes three Key components: Mamba block for long-range feature extraction, CS Mamba and cross modal mamba for shallow and deep feature fusion.}
    \label{fig:mainfig}
\end{figure}
The architecture of our model is depicted in Figure~\ref{fig:mainfig}, comprising three core components: the Mamba block, the channel swapping Mamba block, and the cross modal Mamba block. The Mamba block is instrumental in modeling the long-range dependencies within PAN and LRMS features, while the channel swapping Mamba block and cross-modal Mamba block are employed to explore the relationship between the two modalities. Given the input LRMS and PAN images denoted as $M$ and $P$, the network pipeline can be expressed as follows:

Firstly, we employ convolutional layers to project the two images into the feature space and flatten them along the spatial dimension into tokens:
\begin{align}
&\mathbf{M}_f, \mathbf{P}_f = \phi(\mathbf{M}), \phi(\mathbf{P}),\\
&\mathbf{T}^{ms}_{0}, \mathbf{T}^{pan}_{0} = \text{flatten}(\mathbf{M}_f), \text{flatten}(\mathbf{P}_f).
\end{align}
Subsequently, $\mathbf{T}^{ms}_{0}$ and $\mathbf{T}^{pan}_{0}$ are fed independently into a sequence of Mamba blocks for global feature extraction:
\begin{align}
&\mathbf{T}^{ms}_{i} = \psi_{m_i}..( \psi_{m_1}(\psi_{m_0}(\mathbf{T}^{ms}_{0}))),\
&\mathbf{T}^{pan}_{i} = \psi_{p_i}..( \psi_{p_1}(\psi_{p_0}(\mathbf{T}^{pan}_{0}))).
\end{align}
Here, $\psi_{m_i}(.)$ and $\psi_{p_i}(.)$ denote the i-th Mamba block for extracting LRMS and PAN features, respectively.

Upon obtaining the global features $\mathbf{T}^{ms}_{i}$ and $\mathbf{T}^{pan}_{i}$, we leverage the channel swapping Mamba to enhance feature interaction and obtain $\mathbf{T}^{ms}_{k}$ and $\mathbf{T}^{pan}_{k}$. Subsequently, we utilize the cross modal Mamba block for deep feature fusion. After that, we reshape the MS token into the spatial dimension. The final output is obtained through convolution layers and residual connections:
\begin{align}
&\mathbf{T}^{ms}_{k+j} = \theta_{j}..( \theta_{1}(\theta_{0}(\mathbf{T}^{ms}_{k}, \mathbf{T}^{pan}_{k}), \mathbf{T}^{pan}_{k}), \mathbf{T}^{pan}_{k}),\\
&\mathbf{M}_{\text{out}} = \phi(\text{reshape}(\mathbf{T}^{ms}_{k+j})) + \mathbf{M}.
\end{align}
Here, $\theta(\cdot)$ represents the cross modal Mamba block, and $\phi(\cdot)$ denotes the convolution layers for channel adjustment.

\subsection{Key Components}
\vspace{-0.5em}
\subsubsection{Mamba Block}
Motivated by the Mamba, we employ the Mamba block to extract features and model long-range dependencies. A comprehensive overview of the operations is provided in Algorithm~\ref{Mambablock}. Specifically, the input token sequence $\mathbf{T}_{l-1} \in \mathbf{R^{B\times N \times C}}$ undergoes initial normalization via layer normalization. Subsequently, the normalized sequence is projected into $\mathbf{x} \in \mathbf{R^{B\times N \times P}}$ and $\mathbf{z} \in \mathbf{R^{B\times N \times P}}$ using a multi-layer perceptron (MLP). Following this, a 1-D convolution layer with SiLU activation is applied to process $\mathbf{x}$ and yield $\mathbf{x}'$. Further projection of $\mathbf{x}'$ onto $\mathbf{A}$, $\mathbf{B}$, and $\mathbf{C}$ is performed, and the timescale parameter $\mathbf{\Delta}$ is employed to convert them into discrete versions $\mathbf{\Bar{A}}$ and $\mathbf{\Bar{B}}$. The parameter generation process is delineated in Algorithm~\ref{pf} and corresponds to formula~\ref{e3}. After that, the output $\mathbf{y}$ is computed through the SSM. Subsequently, $\mathbf{y}$ is gated by $\mathbf{z}$ and added to the input $\mathbf{T}_{l-1}$, to get the output sequence $\mathbf{T}_l$. The SSM process is describe in formula~\ref{e7}. 
The computational complexity of Mamba block exhibits linearity to the sequence length N. Specifically, the computational complexity is expressed as $3N(2D)C + N(2D)C$. $D$ and $C$ are fixed numbers.
\begin{minipage}{\textwidth}
\begin{algorithm}[H]
\caption{Parameters Function\label{pf}}
\small
\begin{algorithmic}[1]
\REQUIRE{token sequence $\mathbf{x}'$ : \textcolor{shapecolor}{$(\mathtt{B}, \mathtt{N}, \mathtt{P})$}}
\ENSURE{token sequence $\mathbf{\Bar{A}}$ : \textcolor{shapecolor}{$(\mathtt{B}, \mathtt{N}, \mathtt{P}, \mathtt{K})$},$\mathbf{\Bar{B}}$ : \textcolor{shapecolor}{$(\mathtt{B}, \mathtt{N}, \mathtt{P}, \mathtt{K})$},$\mathbf{\Bar{C}}$ : \textcolor{shapecolor}{$(\mathtt{B}, \mathtt{N}, \mathtt{P})$}}
    \STATE $\mathbf{B}$ : \textcolor{shapecolor}{$(\mathtt{B}, \mathtt{N}, \mathtt{K})$} $\leftarrow$ $\mathbf{Linear}^{\mathbf{B}}(\mathbf{x}')$
    \STATE $\mathbf{C}$ : \textcolor{shapecolor}{$(\mathtt{B}, \mathtt{N}, \mathtt{K})$} $\leftarrow$ $\mathbf{Linear}^{\mathbf{C}}(\mathbf{x}')$
    \STATE $\mathbf{\Delta}$ : \textcolor{shapecolor}{$(\mathtt{B}, \mathtt{N}, \mathtt{P})$} $\leftarrow$ $\log(1 + \exp(\mathbf{Linear}^{\mathbf{\Delta}}(\mathbf{x}') + \mathbf{Parameter}^{\mathbf{\Delta}}))$
    \STATE \textcolor{gray}{\text{/* $\mathbf{Parameter}^{\mathbf{A}} \in$  \textcolor{shapecolor}{$R^{(\mathtt{P}, \mathtt{K})}$} */}}
    \STATE $\overline{\mathbf{A}}$ : \textcolor{shapecolor}{$(\mathtt{B}, \mathtt{N}, \mathtt{P}, \mathtt{K})$} $\leftarrow$ $\mathbf{\Delta}_o \bigotimes \mathbf{Parameter}^{\mathbf{A}}$ 
    \STATE $\overline{\mathbf{B}}$ : \textcolor{shapecolor}{$(\mathtt{B}, \mathtt{N}, \mathtt{P}, \mathtt{K})$} $\leftarrow$ $\mathbf{\Delta}_o \bigotimes \mathbf{B}$
    
    Return: $\mathbf{\Bar{A}},\mathbf{\Bar{B}},\mathbf{C}$ 
\label{alg:block}
\end{algorithmic}
\end{algorithm}
\vspace{-4.5em}
\begin{algorithm}[H]
\caption{Mamba Block\label{Mambablock}}
\small
\begin{algorithmic}[1]
\REQUIRE{token sequence $\mathbf{T}_{l-1}$ : \textcolor{shapecolor}{$(\mathtt{B}, \mathtt{N}, \mathtt{C})$}}
\ENSURE{token sequence $\mathbf{T}_{l}$ : \textcolor{shapecolor}{$(\mathtt{B}, \mathtt{N}, \mathtt{C})$}}
\STATE \textcolor{gray}{\text{/* Apply layer normalization to the input sequence $\mathbf{T}_{l-1}$ */}}
\STATE $\mathbf{T}_{l-1}'$ : \textcolor{shapecolor}{$(\mathtt{B}, \mathtt{N}, \mathtt{C})$} $\leftarrow$ $\mathbf{Norm}(\mathbf{T}_{l-1})$
\STATE $\mathbf{x}$ : \textcolor{shapecolor}{$(\mathtt{B}, \mathtt{N}, \mathtt{P})$} $\leftarrow$ $\mathbf{Linear}^\mathbf{x}(\mathbf{T}_{l-1}')$
\STATE $\mathbf{z}$ : \textcolor{shapecolor}{$(\mathtt{B}, \mathtt{N}, \mathtt{P})$} $\leftarrow$ $\mathbf{Linear}^\mathbf{z}(\mathbf{T}_{l-1}')$
\STATE \textcolor{gray}{\text{/* process the input sequence */}}
\STATE $\mathbf{x}'$ : \textcolor{shapecolor}{$(\mathtt{B}, \mathtt{N}, \mathtt{P})$} $\leftarrow$ $\mathbf{SiLU}(\mathbf{Conv1d}(\mathbf{x}))$
\STATE $\overline{\mathbf{A}}$:\textcolor{shapecolor}{$(\mathtt{B}, \mathtt{N}, \mathtt{P},\mathtt{K})$}  , 
$\overline{\mathbf{B}}$:\textcolor{shapecolor}{$(\mathtt{B}, \mathtt{N}, \mathtt{P},\mathtt{K})$}, $\mathbf{C}$ :\textcolor{shapecolor}{$(\mathtt{B}, \mathtt{N},\mathtt{K})$}$\leftarrow$  $\mathbf{Parameters Function}$($\mathbf{x}'$)
    \STATE $\mathbf{y}$ : \textcolor{shapecolor}{$(\mathtt{B}, \mathtt{N}, \mathtt{P})$} $\leftarrow$ $\mathbf{SSM}(\overline{\mathbf{A}}, 
    \overline{\mathbf{B}}, \mathbf{C})(\mathbf{x}')$

\STATE \textcolor{gray}{\text{/* Obtain the gated $\mathbf{y}$ by multiplying it with a gating factor. */}}
\STATE $\mathbf{y}'$ : \textcolor{shapecolor}{$(\mathtt{B}, \mathtt{N}, \mathtt{P})$} $\leftarrow$ $\mathbf{y}\bigodot \mathbf{SiLU}(\mathbf{z}) $
\STATE \textcolor{gray}{\text{/* residual connection */}}
\STATE $\mathbf{T}_{l}$ : \textcolor{shapecolor}{$(\mathtt{B}, \mathtt{N}, \mathtt{C})$} $\leftarrow$ $\mathbf{Linear}^\mathbf{T}(\mathbf{y}') + \mathbf{T}_{l-1}$

Return: $\mathbf{T}_{l}$ 
\label{alg:block}
\end{algorithmic}
\end{algorithm}
\end{minipage}
\subsubsection{Channel Swapping Mamba Block}
To encourage feature interaction between PAN and LRMS modalities and initiate a correlation between them, we introduced a Mamba fusion block based on channel swapping. This module efficiently swaps channels between LRMS and PAN features, facilitating lightweight feature interaction. The swapped features are then processed through the Mamba block. The channel swapping operation enhances cross-modal correlations by incorporating information from distinct channels, thereby enriching the diversity of channel features, which contributes to the overall improvement in model performance.
Given LRMS features $\mathbf{T}^{ms}_{l-1} \in \mathbf{R}^{B,N,C}$ and PAN features $\mathbf{T}^{pan}_{l-1} \in \mathbf{R}^{B,N,C}$ as inputs, we split each feature along the channel dimension into two equal portions. 
 The first half of channels from $\mathbf{T}^{ms}_{l-1}$ is concatenated with the latter half of $\mathbf{T}^{pan}_{l-1}$ and processed through the Mamba block for feature extraction. The obtained features are added to $\mathbf{T}^{ms}_{l-1}$, resulting in the creation of a new feature $\mathbf{T}^{ms}_{l}$. Simultaneously, the first half of $\mathbf{T}^{pan}_{l-1}$ is concatenated with the latter half of $\mathbf{T}^{ms}_{l-1}$ and passed through the Mamba block. The resulting features are added to $\mathbf{T}^{pan}_{l-1}$, generating $\mathbf{T}^{pan}_{l}$.
 These features encapsulate information from both modalities, enhancing overall feature diversity.
 \begin{algorithm}[h]
\caption{Cross Modal Mamba Block \label{cmmb}}
\small
\begin{algorithmic}[1]
\REQUIRE{token sequence $\mathbf{T}^{ms}_{l-1}$ : \textcolor{shapecolor}{$(\mathtt{B}, \mathtt{N}, \mathtt{C})$},$\mathbf{T}^{pan}_{l-1}$ : \textcolor{shapecolor}{$(\mathtt{B}, \mathtt{N}, \mathtt{C})$}}
\ENSURE{token sequence $\mathbf{T}^{ms}_{l}$ : \textcolor{shapecolor}{$(\mathtt{B}, \mathtt{N}, \mathtt{C})$}}
\FOR{$o$ in \{pan, ms\}}
\STATE \textcolor{gray}{\text{/* Apply layer normalization to the input sequence $\mathbf{T}_{l-1}$ */}}
\STATE $\mathbf{T{'}}^{o}_{l-1}$ : \textcolor{shapecolor}{$(\mathtt{B}, \mathtt{N}, \mathtt{C})$} $\leftarrow$ $\mathbf{Norm}(\mathbf{T}^{o}_{l-1})$
\STATE $\mathbf{x}_o$ : \textcolor{shapecolor}{$(\mathtt{B}, \mathtt{N}, \mathtt{P})$} $\leftarrow$ $\mathbf{Linear}^\mathbf{x}_o(\mathbf{T{'}}^{o}_{l-1})$
\STATE \textcolor{gray}{\text{/* process with sequence */}}
\STATE $\mathbf{x}'_o$ : \textcolor{shapecolor}{$(\mathtt{B}, \mathtt{N}, \mathtt{P})$} $\leftarrow$ $\mathbf{SiLU}(\mathbf{Conv1d}_o(\mathbf{x}_o))$
\STATE $\overline{\mathbf{A}}_o$:\textcolor{shapecolor}{$(\mathtt{B}, \mathtt{N}, \mathtt{P},\mathtt{K})$}  , 
$\overline{\mathbf{B}}_o$:\textcolor{shapecolor}{$(\mathtt{B}, \mathtt{N}, \mathtt{P},\mathtt{K})$}, $\mathbf{C}_o$ :\textcolor{shapecolor}{$(\mathtt{B}, \mathtt{N},\mathtt{K})$}$\leftarrow$  $\mathbf{Parameters Function_o}$($\mathbf{x}'_o$)
    \STATE $\mathbf{y}_o$ : \textcolor{shapecolor}{$(\mathtt{B}, \mathtt{N}, \mathtt{P})$} $\leftarrow$ $\mathbf{SSM}(\overline{\mathbf{A}}_o, 
    \overline{\mathbf{B}}_o, \mathbf{C}_o)(\mathbf{x}'_o)$
\ENDFOR

\STATE $\mathbf{z}$ : \textcolor{shapecolor}{$(\mathtt{B}, \mathtt{N}, \mathtt{P})$} $\leftarrow$ $\mathbf{Linear}^\mathbf{z}(\mathbf{T{'}}^{ms}_{l-1})$
\STATE \textcolor{gray}{\text{/* Obtain the gated $\mathbf{y}$ by multiplying it with a gating factor. */}}
\STATE $\mathbf{y}_{ms}'$ : \textcolor{shapecolor}{$(\mathtt{B}, \mathtt{N}, \mathtt{P})$} $\leftarrow$ $\mathbf{y}_{ms} \bigodot \mathbf{SiLU}(\mathbf{z}) $
\STATE $\mathbf{y}_{pan}'$ : \textcolor{shapecolor}{$(\mathtt{B}, \mathtt{N}, \mathtt{P})$} $\leftarrow$ $\mathbf{y}_{pan} \bigodot \mathbf{SiLU}(\mathbf{z}) $
\STATE \textcolor{gray}{\text{/* residual connection */}}
\STATE $\mathbf{F}^{ms}_{l-1}$ : \textcolor{shapecolor}{$(\mathtt{B}, \mathtt{C}, \mathtt{H},\mathtt{W})$} $\leftarrow$ $\mathbf{Reshape(\mathbf{Linear}^\mathbf{T}(\mathbf{y}_{ms}' + \mathbf{y}_{pan}') + \mathbf{T}^{ms}_{l-1})}$
\STATE \textcolor{gray}{\text{/* depth-wise convolution */}}
\STATE $\mathbf{T}^{ms}_{l}$: \textcolor{shapecolor}{$(\mathtt{B},\mathtt{N},\mathtt{C})$}
$\leftarrow$ $\mathbf{Flatten}(\mathbf{DWConv}(\mathbf{F}^{ms}_{l-1})+\mathbf{F}^{ms}_{l-1})$

Return: $\mathbf{T}^{ms}_{l}$ 
\label{alg:block}
\end{algorithmic}
\end{algorithm}
\subsubsection{Cross modality Mamba Block}
Motivated by the concept of Cross Attention~\cite{chen2021crossvit}, we introduce a novel cross modal Mamba block designed for facilitating cross-modal feature interaction and fusion. In this approach, we project features from two modalities into a shared space, employing gating mechanisms to encourage complementary feature learning while suppressing redundant features. Concurrently, to enhance local features, we incorporate depth-wise convolution within the module, thereby amplifying the encoding capacity of local features during the fusion process. The details of this module are elucidated in Algo~\ref{cmmb}.
The generation of $y_{ms}$ and $y_{pan}$ follows the process outlined in the Mamba block. Subsequently, we obtain the gating parameters $z$ by projecting $\mathbf{T{'}}^{ms}_{l-1} $ and employ $z$ to modulate $y_{ms}$ and $y_{pan}$. The fusion of these two features involves addition, followed by reshaping to obtain a 2-D feature $\mathbf{F}^{ms}_{l-1}$. To enhance locality, we apply depth-wise convolution, and subsequently flatten the feature to a 1-D sequence, generating output sequence $\mathbf{T}^{ms}_{l}$.

\subsection{Loss Function}
In alignment with the prevalent practices in this filed, we adopt the L1 loss as our chosen loss function. Specifically, with the output denoted as $\mathbf{M_{out}}$ and the corresponding ground truth as $\mathbf{G}$, the loss function is expressed as:
\begin{align}
    \mathcal{L} = ||\mathbf{G}-\mathbf{M_{out}}||_1
\end{align}

\section{Experiment}
\subsection{Datasets and Benchmark}
In our experiments, we selected datasets comprising WorldView-II (WV2), Gaofen-2 (GF2) and WorldView-III (WV3), characterized by diverse resolutions and a broad spectrum of scenes. Specifically, WorldView-II encompasses industrial areas and natural landscapes, Gaofen 2 encompasses mountains and rivers, while WorldView-III predominantly features urban roads and urban scenes. Given the absence of ground truth, the dataset generating process adheres to the Wald protocol~\cite{gt}. For comparative analysis, we opted for a selection of representative traditional methods, including GFPCA~\cite{GFPCA}, GS~\cite{GS}, Brovey~\cite{Brovey}, IHS~\cite{IHS}, and SFIM~\cite{SFIM}, alongside advanced deep learning methods such as PanNet~\cite{yang2017pannet}, MSDCNN~\cite{msdcnn}, SRPPNN~\cite{srppnn}, INNformer~\cite{zhou2022pan}, SFINet~\cite{zhou2022spatial}, MSDDN~\cite{he2023multi}, and PanFlowNet~\cite{yang2023panflownet}. The chosen evaluation metrics encompass PSNR, SSIM, SAM (spectral angle mapper)~\cite{Yuhas1992DiscriminationAS}, ERGAS (relative global error in synthesis)~\cite{ergas} and Non-reference metrics including $D_s$, $D_\lambda$ and QNR.

\subsection{Implement Details}
Utilizing the PyTorch framework, our code implementation and training procedures are executed on an Nvidia V100 GPU. The model features are configured with N=32 channels.

We initialize the learning rate at 5e-4, employing a cosine decay scheduling strategy. After 500 epochs, the learning rate diminishes to 5e-8. Optimization is carried out using the Adam optimizer, with gradient clipping set to 4 for training stability. Given variations in data sizes, we set the number of training epochs to 200 for the WorldView-II dataset and 500 for the Gaofen-2 and WorldView-III datasets.

\subsection{Comparison with State of Arts Methods}
\begin{table}[h]
\centering
\normalsize
\renewcommand{\arraystretch}{1.2}
\renewcommand{\tabcolsep}{3pt}
\caption{Quantitative comparison on three datasets. Best results highlighted in \textcolor{red}{red}. $\uparrow$ indicates that the larger the value, the
better the performance, and $\downarrow$ indicates that the smaller the value, the better the performance.\label{cmp}}
\resizebox{\linewidth}{!}
{
\begin{tabular}{c|cccc|cccc|cccc}
\hline
                         & \multicolumn{4}{c|}{WorldView-II}                                                                                              & \multicolumn{4}{c|}{Gaofen-2}                                                                                                  & \multicolumn{4}{c}{WorldView-III}                                                                                              \\ \cline{2-13} 
\multirow{-2}{*}{Method} & PSNR$\uparrow$                           & SSIM$\uparrow$                          & SAM$\downarrow$                           & ERGAS$\downarrow$                         & PSNR$\uparrow$                           & SSIM$\uparrow$                          & SAM$\downarrow$                           & ERGAS$\downarrow$                         & PSNR$\uparrow$                           & SSIM $\uparrow$                         & SAM$\downarrow$                           & ERGAS$\downarrow$                         \\ \hline
SFIM                     & 34.1297                        & 0.8975                        & 0.0439                        & 2.3449                        & 36.906                         & 0.8882                        & 0.0318                        & 1.7398                        & 21.8212                        & 0.5457                        & 0.1208                        & 8.9730                        \\
Brovey                   & 35.8646                        & 0.9216                        & 0.0403                        & 1.8238                        & 37.226                         & 0.9034                        & 0.0309                        & 1.6736                        & 22.5060                        & 0.5466                        & 0.1159                        & 8.2331                        \\
IHS                      & 35.6376                        & 0.9176                        & 0.0423                        & 1.8774                        & 37.7974                        & 0.9026                        & 0.0218                        & 1.3720                        & 22.5608                        & 0.5470                        & 0.1217                        & 8.2433                        \\
GS                       & 35.2962                        & 0.9027                        & 0.0461                        & 2.0278                        & 38.1754                        & 0.9100                        & 0.0243                        & 1.5336                        & 22.5579                        & 0.5354                        & 0.1266                        & 8.3616                        \\
GFPCA                    & 34.5581                        & 0.9038                        & 0.0488                        & 2.1411                        & 37.9443                        & 0.9204                        & 0.0314                        & 1.5604                        & 22.3344                        & 0.4826                        & 0.1294                        & 8.3964                        \\ \hline
PanNet                   & 40.8176                        & 0.9626                        & 0.0257                        & 1.0557                        & 43.0659                        & 0.9685                        & 0.0178                        & 0.8577                        & 29.6840                        & 0.9072                        & 0.0851                        & 3.4263                        \\
MSDCNN                   & 41.3355                        & 0.9664                        & 0.0242                        & 0.9940                        & 45.6847                        & 0.9827                        & 0.0135                        & 0.6389                        & 30.3038                        & 0.9184                        & 0.0782                        & 3.1884                        \\
SRPPNN                   & 41.4538                        & 0.9679                        & 0.0233                        & 0.9899                        & 47.1998                        & 0.9877                        & 0.0106                        & 0.5586                        & 30.4346                        & 0.9202                        & 0.0770                        & 3.1553                        \\
INNformer                & 41.6903                        & 0.9704                        & 0.0227                        & 0.9514                        & 47.3528                        & 0.9893                        & 0.0102                        & 0.5479                        & 30.5365                        & 0.9225                        & 0.0747                        & 3.1142                        \\
SFINet                   & 41.7244                        & 0.9725                        & 0.0220                        & 0.9506                        & 47.4712                        & \textcolor{red}{0.9901}                        & 0.0102                        & 0.5479                        & 30.5901                        & 0.9236                        & 0.0741                        & 3.0798                        \\
MSDDN                    & 41.8435                        & 0.9711                        & 0.0222                        & 0.9478                        & 47.4101                        & { 0.9895} & {\color[HTML]{FE0000} 0.0101} & 0.5414                        & 30.8645                        & 0.9258                        & 0.0757                        & 2.9581                        \\
PanFlowNet               & 41.8548                        & 0.9712                        & 0.0224                        & 0.9335                        & 47.2533                        & 0.9884                        & 0.0103                        & 0.5512                        & 30.4873                        & 0.9221                        & 0.0751                        & 2.9531                        \\ \hline
Ours                     & {\color[HTML]{FE0000} 42.2354} & {\color[HTML]{FE0000} 0.9729} & {\color[HTML]{FE0000} 0.0212} & {\color[HTML]{FE0000} 0.8975} & {\color[HTML]{FE0000} 47.6453} & 0.9894                        & 0.0103                        & {\color[HTML]{FE0000} 0.5286} & {\color[HTML]{FE0000} 31.1551} & {\color[HTML]{FE0000} 0.9299} & {\color[HTML]{FE0000} 0.0702} & {\color[HTML]{FE0000} 2.8942} \\ \hline
\end{tabular}

}
\end{table}

\begin{table}[!h]
\normalsize
\centering
{\caption{\label{fwv2}
Evaluation of the proposed method on real-world full-resolution scenes from the WorldView-II dataset.}}
\resizebox{\linewidth}{!}
{
\begin{tabular}{c|ccccccccccc}
\hline
Metric &SFIM & Brovey & IHS& GFPCA& MSDCNN & SRPPNN & INNformer &SFINet &PanFlowNet &

Ours \\
\hline
$D_{\lambda}$ $\downarrow$ & 0.1403 & 0.1026 & 0.1110 & 0.1139  & 0.1063 & 0.0998 &0.0995 &0.1034&0.0966&\textcolor{red}{{0.0966}} \\
$D_{S}$ $\downarrow$  & 0.1320 & 0.1409 & 0.1556 & 0.1535 & 0.1443 &  0.1637&0.1305&0.1305&0.1274&\textcolor{red}{0.1272} \\
QNR $\uparrow$ & 0.7826 & 0.7728 & 0.7527 & 0.7532 &  0.7683 & 0.7548 &0.7858 &0.7827&0.7910&\textcolor{red}{0.7911} \\ \hline
\end{tabular}}
\end{table}

\subsubsection{Evaluation on Reduced-Resolution Scene}
In our comparative analysis depicted in Table~\ref{cmp}, we benchmarked our proposed method against state-of-the-art methods in the field. The results highlight significant improvement achieved by our proposed network structure, outperforming other methods across various evaluation metrics. Notably, on both the WorldView-II and WorldView-III datasets, our method demonstrated noteworthy improvements in PSNR metrics, with enhancements of 0.38 and 0.29, respectively. This indicates a closer alignment of our results with the ground truth. Similar trends are observed in the SSIM indicator, while the SAM indicator signifies spectral similarity. Our spectral similarity on WV2 surpasses state-of-the-art methods, with comparable outcomes on the WV3 and GF2 dataset. The ERGAS indicator validate the overall superior performance of our method across each spectral band, substantiating its efficacy.

\begin{figure}[!h]
    \centering
    \includegraphics[width=\textwidth]{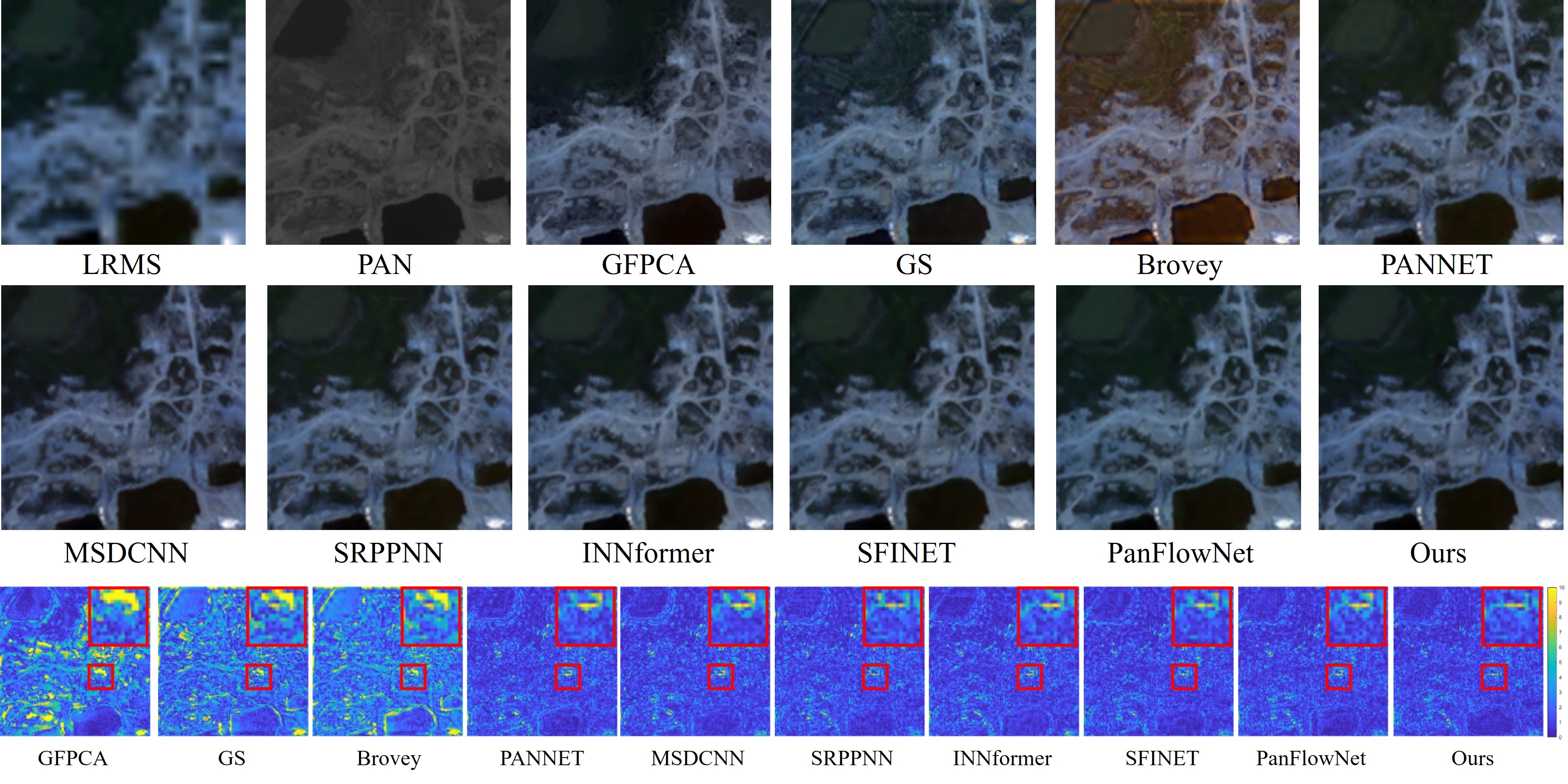}
    \caption{The result of our approach was compared against nine other methods on WorldView-II dataset.}
    \label{fig:wv2}
\end{figure}

\begin{figure}[!h]
    \centering
    \includegraphics[width=\textwidth]{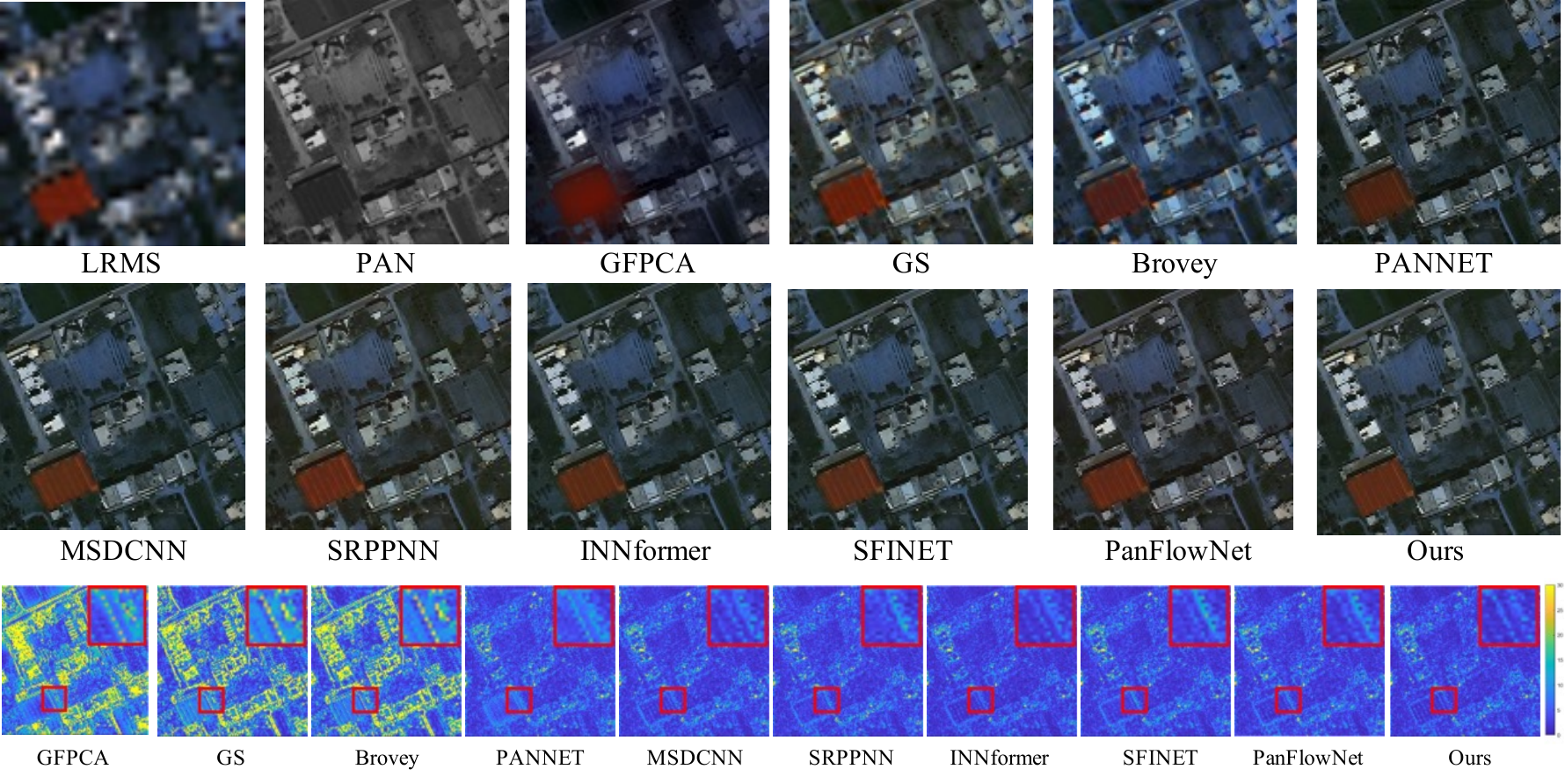}
    \caption{The result of our approach was compared against nine other methods on WorldView-III dataset.}
    \label{fig:wv3}
\end{figure}

Quantitative results of the model are illustrated in Figures~\ref{fig:wv2} and~\ref{fig:wv3}. Representative samples were chosen for visualization, and the last row of each image depicts the mean square error map of the pan-sharpened result in comparison to the ground truth. Brighter regions indicate greater discrepancies. Our method consistently yields the smallest errors, underscoring its proximity to the ground truth. Furthermore, upon pan-sharpened the results, our method exhibits superior performance in extracting high-frequency information and preserving spectral details, leading to clearer texture.
\subsubsection{Evaluation on Full-Resolution Scene}
\begin{figure}[!h]
    \centering
    \includegraphics[width=\textwidth]{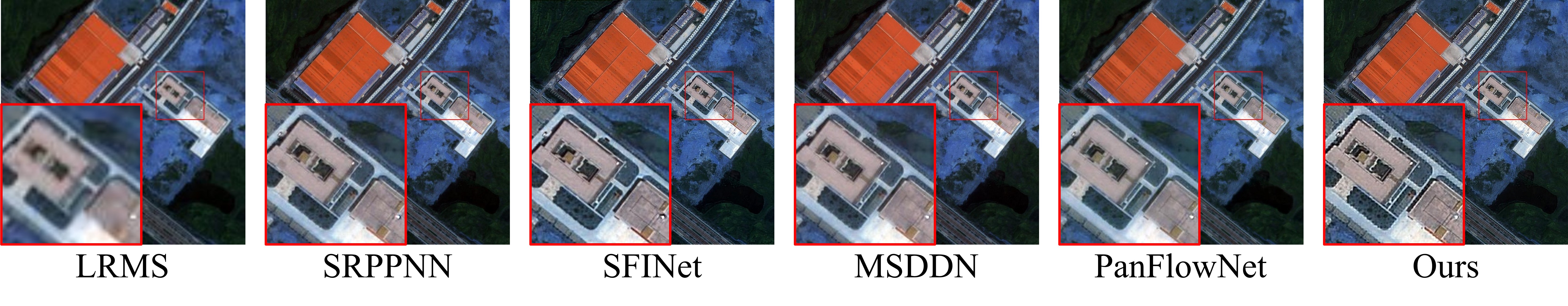}
    \caption{The result of our approach was compared against four other methods on full-resolution WV2 dataset.}
    \label{fig:fwv2}
\end{figure}

To validate our model's performance in real-world settings, we conducted evaluations using the full-resolution WV2 dataset. This dataset was utilized in its original form without downsampling, presenting a scenario with consistent real-world degradation. Owing to the absence of ground truth, we employed non-reference quality metrics—namely, $D_s$, $D_\lambda$, and QNR—for assessing model performance. As delineated in Table~\ref{fwv2}, our approach outperforms the current SOTA across all three evaluated metrics, thereby affirming our model's robust generalizability in real-world contexts. Additionally, we have provided visual results of our method applied to the full-resolution WV2 dataset in the Figure~\ref{fig:fwv2}, showcasing its superior image fusion capabilities.


\begin{table}[h]
\caption{Ablation for Pan-Mamba on Worldview-II datasets. The PSNR/SSIM/SAM/ERGAS values on benchmarks are reported. "A → B" is to replace A with B. "None"  The numbers of parameters, \#FLOPs  are counted by the thops library with a resolution of 128 × 128 pixels. Best results are highlighted in \textcolor{red}{red}.}
\label{abl}
\centering
\normalsize
 	\renewcommand{\tabcolsep}{3pt} 
\renewcommand{\arraystretch}{1.5}
\resizebox{\linewidth}{!}{
\begin{tabular}{|c|c|c|c|c|}
\hline
Ablation                        & Variant                                   & \#Params {[}M{]} & \#FLOPs {[}G{]} & WorldView-II                  \\ \hline
Baseline                        & -                                         & 0.1827           & 3.0028          & \textcolor{red}{42.2354}/\textcolor{red}{0.9729}/\textcolor{red}{0.0212}/0.8975  \\ \hline
\multirow{4}{*}{Core Operation} & Mamba-\textgreater{}Conv                  & 0.6235           & 10.1879         & 42.0059/0.9722/0.0214/0.9208 \\ \cline{2-5} 
                                & Mamba-\textgreater{}Transposed Attention  & 0.4944           & 8.0861          & 42.1568/0.9726/0.0212/0.9016  \\ \cline{2-5} 
                                & Mamba-\textgreater{}Window Attention      & 0.4720           & 7.6913          & 42.0980/0.9721/0.0212/0.9041  \\ \cline{2-5} 
                                & Mamba-\textgreater{}Self Attention        & 3.2424           & 14.9305         & 36.4403/0.9096/0.0303/1.5648  \\ \hline
\multirow{3}{*}{Main Module}    & Mamba Block-\textgreater{}Conv3x3         & 0.3307           & 5.4248          & 42.1659/0.9725/0.0211/0.8979  \\ \cline{2-5} 
                                & Channel Swapping Mamba-\textgreater{}None & 0.1827           & 3.0008          & 42.2294/0.9727/0.0212/\textcolor{red}{0.8972}  \\ \cline{2-5} 
                                & Cross Mamba-\textgreater{}None            & 0.1811           & 2.9852          & 42.0733/0.9716/0.0215/0.9109  \\ \hline
\end{tabular}

}
\end{table}

\subsection{Ablation Study}
To comprehensively assess the contribution of each module, we conducted ablation experiments. To ensure fairness in comparison, we conducted training and reported experimental results exclusively on the WorldView-II dataset, maintaining consistent experimental configurations across all ablation experiments. Our ablation experiment consists of two main parts. The first part compare the network's core operator with prevalent operators in low-level vision. The second part validates the proposed three modules by selectively removing or substituting each module to assess its individual functionality.

\subsubsection{Effectiveness of the State Space Model.}
The ablation experiment conducted in the first part aims to validate the performance of our network's core operator, the state space model, as illustrated in rows two to four of Table~\ref{abl}. During this experiment, we substituted the state space model with four widely recognized operators in the image restoration domain: the convolution operation from Hinet~\cite{chen2021hinet}, transposed attention from restormer~\cite{zamir2022restormer}, window attention from swinir~\cite{liang2021swinir}, and self-attention from IPT~\cite{chen2021pre}. The feature extraction module was constructed directly employing the aforementioned operators. In the fusion module, Convolution layers is a hybrid of Hinet and convolution layer with $1\times1$ kernel, while other types utilize their respective cross-attention variants. Consistent hyperparameter settings were applied across all experiments to ensure a fair comparison.
Upon these experiments, it becomes evident that the mamba module's core operation, namely the state space model, boasts the fewest parameters and requires the least computational resources. Moreover, it exhibits superior performance on the WV2 dataset. It is noteworthy that self-attention, while a powerful mechanism, fails to achieve optimal results on small-scale remote sensing datasets, attributable to its inherent lack of inductive bias.
\subsubsection{Effectiveness of the Mamba Block.}
This ablation experiment is devised to validate the efficacy of the Mamba block in feature extraction. Within our model, the Mamba block plays a crucial role in modeling long-range dependencies in features. In this set of experiments, we substituted the Mamba block with a standard $3 \times 3$ convolution operation and relocated the flattening operation in the model post feature extraction. The results from the sixth row of table~\ref{abl} reveal a decline in model performance when the Mamba block is removed, providing evidence of the block's effectiveness.
\subsubsection{Effectiveness of the channel swapping Mamba Block.}
The second set of ablation experiments in second part is designed to validate the effectiveness of the channel swapping Mamba. Within our architecture, the channel swapping Mamba is specifically designed for shadow feature fusion, enhancing the diversity of channel features. In this experiment, we eliminated the channel swapping operation along with its associated Mamba block. As shown in the seventh row of Table~\ref{abl}, despite the low computational cost of this module, the removal resulted in a noticeable decline in performance, underscoring the efficacy of this module.
\subsubsection{Effectiveness of the cross mmodal Mamba Block.}
The third set of ablation experiments in second part is conducted to substantiate the effectiveness of our pivotal fusion module, the cross-modal Mamba block. This module is designed to conduct deep fusion on LRMS and PAN features, reducing redundant features through gating mechanisms. In this experiment, we directly remove the cross modal Mamba block and exclusively employed the Channel Swapping Mamba for feature fusion. The results depicted in the Table~\ref{abl} showcase a substantial decline in the model's performance following the removal of the cross-modal Mamba block.
\begin{figure}
    \centering
    \includegraphics[width=\textwidth]{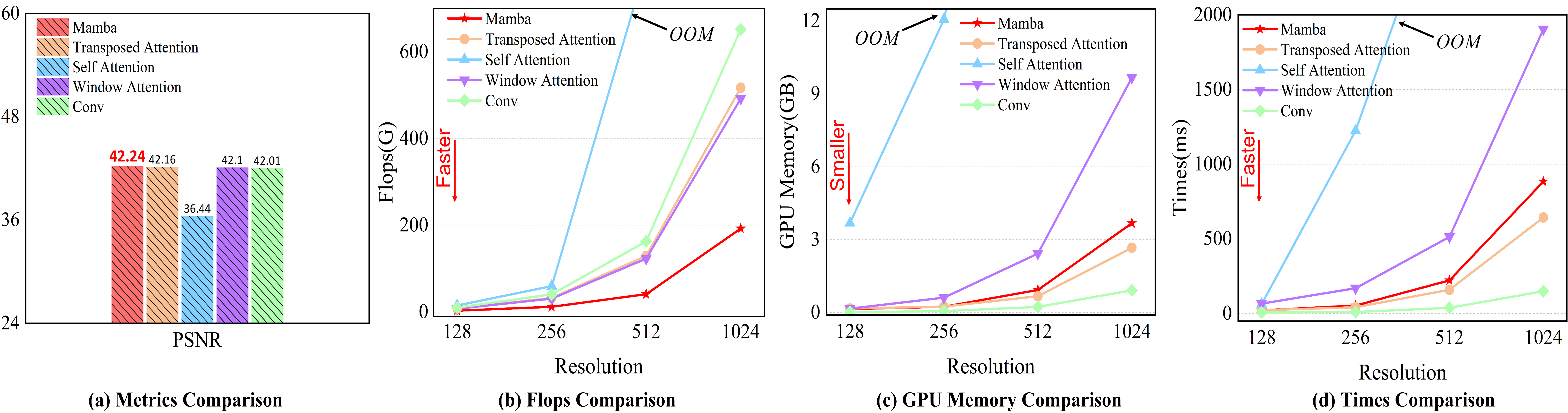}
    \caption{Performance and efficiency comparisons between different operators in ablation section and our model.}
    \label{fig:histgra}
\end{figure}
\begin{table}[h]
    \centering
    \renewcommand{\arraystretch}{1.2}
\renewcommand{\tabcolsep}{3pt}
\caption{Comparisons on flops and parameter numbers.}
    \label{tab:flp}
    \resizebox{\linewidth}{!}{
\begin{tabular}{c|cccccccc}
\hline
Methods   & PanNet & MSDCNN & SRPPNN  & INNformer & SFINet & MSDDN  & PanFlowNet & Ours   \\ \hline
Params(M) & 0.0688 & 0.2390 & 1.7114  & 0.0706    & 0.0871 & 0.3185 & 0.0873     & 0.1827 \\ \hline
FLOPS(G)  & 1.1275 & 3.9158 & 21.1059 & 1.3079    & 1.2558 & 2.5085 & 5.7126     & 3.0088 \\ \hline
\end{tabular}
}
\end{table}
\subsection{Comparison of Efficiency}
We conducted an efficiency analysis of our model, which comprises two main parts. The first part involves comparing the computational complexity between our method and several operators mentioned in the ablation study, while the second part focuses on comparing the computational complexity with the benchmark methods.

In Figure~\ref{fig:histgra}, we assess the complexity of our method in comparison to Self-attention, transposed attention, window attention, and convolution, with the resolution of the input PAN image scales from $128 \times 128$ to $1024 \times 1024$. The results reveal our method's superior performance, with the best flops metric. Regarding GPU memory usage and inference time, our method exhibits complexity comparable to transposed attention, significantly lower than window attention and self-attention. Notably, the convolution operator, despite having low memory requirements, hampers model performance due to its limited local receptive field. Conversely, the linear complexity of transposed attention lacks the ability to capture spatial dependencies, while Mamba blocks effectively model global spatial dependencies with linear complexity.

In the second part of the comparison, detailed in the Table~\ref{tab:flp}, our method is compared with the sota method. With similar parameter and computational complexity, our approach achieves best results, underscoring the capabilities of our model.

\section{Conclusion}
In this study, drawing inspiration from the State Space model, we introduce a novel Pan-sharpening network, termed Pan Mamba. This innovative network incorporates Mamba blocks, channel swapping Mamba blocks, and cross modal Mamba blocks. The proposed network achieves efficient global feature extraction and facilitates cross-modal information exchange with linear complexity. Notably, it outperforms state-of-the-art methods with a lightweight model on publicly available remote sensing datasets, showcasing robust spectral accuracy and adept preservation of texture information.

%
%
\bibliographystyle{splncs04}
\bibliography{main}

\begin{thebibliography}{10}
\providecommand{\url}[1]{\texttt{#1}}
\providecommand{\urlprefix}{URL }
\providecommand{\doi}[1]{https://doi.org/#1}

\bibitem{ergas}
Alparone, L., Wald, L., Chanussot, J., Thomas, C., Gamba, P., Bruce, L.M.: Comparison of pansharpening algorithms: Outcome of the 2006 grs-s data fusion contest. IEEE Transactions on Geoscience and Remote Sensing  \textbf{45}(10),  3012--3021 (2007)

\bibitem{srppnn}
Cai, J., Huang, B.: Super-resolution-guided progressive pansharpening based on a deep convolutional neural network. IEEE Transactions on Geoscience and Remote Sensing  \textbf{59}(6),  5206--5220 (2021). \doi{10.1109/TGRS.2020.3015878}

\bibitem{chen2021crossvit}
Chen, C.F.R., Fan, Q., Panda, R.: Crossvit: Cross-attention multi-scale vision transformer for image classification. In: Proceedings of the IEEE/CVF international conference on computer vision. pp. 357--366 (2021)

\bibitem{chen2021pre}
Chen, H., Wang, Y., Guo, T., Xu, C., Deng, Y., Liu, Z., Ma, S., Xu, C., Xu, C., Gao, W.: Pre-trained image processing transformer. In: Proceedings of the IEEE/CVF conference on computer vision and pattern recognition. pp. 12299--12310 (2021)

\bibitem{chen2021hinet}
Chen, L., Lu, X., Zhang, J., Chu, X., Chen, C.: Hinet: Half instance normalization network for image restoration. In: Proceedings of the IEEE/CVF Conference on Computer Vision and Pattern Recognition. pp. 182--192 (2021)

\bibitem{srcnn}
Dong, C., Loy, C.C., He, K., Tang, X.: Image super-resolution using deep convolutional networks. IEEE Transactions on Pattern Analysis and Machine Intelligence  \textbf{38}(2),  295--307 (2016). \doi{10.1109/TPAMI.2015.2439281}

\bibitem{vit}
Dosovitskiy, A., Beyer, L., Kolesnikov, A., Weissenborn, D., Zhai, X., Unterthiner, T., Dehghani, M., Minderer, M., Heigold, G., Gelly, S., et~al.: An image is worth 16x16 words: Transformers for image recognition at scale. arXiv preprint arXiv:2010.11929  (2020)

\bibitem{fasbender2008bayesian}
Fasbender, D., Radoux, J., Bogaert, P.: Bayesian data fusion for adaptable image pansharpening. IEEE Transactions on Geoscience and Remote Sensing  \textbf{46}(6),  1847--1857 (2008)

\bibitem{Brovey}
Gillespie, A.R., Kahle, A.B., Walker, R.E.: Color enhancement of highly correlated images. ii. channel ratio and "chromaticity" transformation techniques - sciencedirect. Remote Sensing of Environment  \textbf{22}(3),  343--365 (1987)

\bibitem{gu2023mamba}
Gu, A., Dao, T.: Mamba: Linear-time sequence modeling with selective state spaces. arXiv preprint arXiv:2312.00752  (2023)

\bibitem{s4}
Gu, A., Goel, K., R{\'e}, C.: Efficiently modeling long sequences with structured state spaces. arXiv preprint arXiv:2111.00396  (2021)

\bibitem{IHS}
Haydn, R., Dalke, G.W., Henkel, J., Bare, J.E.: Application of the ihs color transform to the processing of multisensor data and image enhancement. National Academy of Sciences of the United States of America  \textbf{79}(13),  571--577 (1982)

\bibitem{he2023multi}
He, X., Yan, K., Zhang, J., Li, R., Xie, C., Zhou, M., Hong, D.: Multi-scale dual-domain guidance network for pan-sharpening. IEEE Transactions on Geoscience and Remote Sensing  (2023)

\bibitem{GS}
Laben, C., Brower, B.: Process for enhancing the spatial resolution of multispectral imagery using pan-sharpening. US Patent 6011875A  (2000)

\bibitem{liang2021swinir}
Liang, J., Cao, J., Sun, G., Zhang, K., Van~Gool, L., Timofte, R.: Swinir: Image restoration using swin transformer. In: Proceedings of the IEEE/CVF international conference on computer vision. pp. 1833--1844 (2021)

\bibitem{GFPCA}
Liao, W., Xin, H., Coillie, F.V., Thoonen, G., Philips, W.: Two-stage fusion of thermal hyperspectral and visible rgb image by pca and guided filter. In: Workshop on Hyperspectral Image and Signal Processing: Evolution in Remote Sensing (2017)

\bibitem{SFIM}
Liu., J.G.: Smoothing filter-based intensity modulation: A spectral preserve image fusion technique for improving spatial details. International Journal of Remote Sensing  \textbf{21}(18),  3461--3472 (2000)

\bibitem{liu2024vmamba}
Liu, Y., Tian, Y., Zhao, Y., Yu, H., Xie, L., Wang, Y., Ye, Q., Liu, Y.: Vmamba: Visual state space model. arXiv preprint arXiv:2401.10166  (2024)

\bibitem{liu2021swin}
Liu, Z., Lin, Y., Cao, Y., Hu, H., Wei, Y., Zhang, Z., Lin, S., Guo, B.: Swin transformer: Hierarchical vision transformer using shifted windows. In: Proceedings of the IEEE/CVF international conference on computer vision. pp. 10012--10022 (2021)

\bibitem{ma2024u}
Ma, J., Li, F., Wang, B.: U-mamba: Enhancing long-range dependency for biomedical image segmentation. arXiv preprint arXiv:2401.04722  (2024)

\bibitem{pnn}
Masi, G., Cozzolino, D., Verdoliva, L., Scarpa, G.: Pansharpening by convolutional neural networks. Remote Sensing  \textbf{8}(7), ~594 (2016)

\bibitem{h3}
Mehta, H., Gupta, A., Cutkosky, A., Neyshabur, B.: Long range language modeling via gated state spaces. arXiv preprint arXiv:2206.13947  (2022)

\bibitem{ATWT1999}
Nunez, J., Otazu, X., Fors, O., Prades, A., Pala, V., Arbiol, R.: Multiresolution-based image fusion with additive wavelet decomposition. IEEE Transactions on Geoscience and Remote sensing  \textbf{37}(3),  1204--1211 (1999)

\bibitem{HPF}
Schowengerdt, R.A.: Reconstruction of multispatial, multispectral image data using spatial frequency content. Photogrammetric Engineering and Remote Sensing  \textbf{46}(10),  1325--1334 (1980)

\bibitem{smith2022simplified}
Smith, J.T., Warrington, A., Linderman, S.W.: Simplified state space layers for sequence modeling. arXiv preprint arXiv:2208.04933  (2022)

\bibitem{gt}
Wald, L., Ranchin, T., Mangolini, M.: Fusion of satellite images of different spatial resolutions: Assessing the quality of resulting images. Photogrammetric Engineering and Remote Sensing  \textbf{63},  691--699 (11 1997)

\bibitem{xiao2023random}
Xiao, J., Fu, X., Zhou, M., Liu, H., Zha, Z.J.: Random shuffle transformer for image restoration. In: International Conference on Machine Learning. pp. 38039--38058. PMLR (2023)

\bibitem{xing2024segmamba}
Xing, Z., Ye, T., Yang, Y., Liu, G., Zhu, L.: Segmamba: Long-range sequential modeling mamba for 3d medical image segmentation. arXiv preprint arXiv:2401.13560  (2024)

\bibitem{gppnn}
Xu, S., Zhang, J., Zhao, Z., Sun, K., Liu, J., Zhang, C.: Deep gradient projection networks for pan-sharpening. In: IEEE Conference on Computer Vision and Pattern Recognition. pp. 1366--1375 (June 2021)

\bibitem{yang2023panflownet}
Yang, G., Cao, X., Xiao, W., Zhou, M., Liu, A., Chen, X., Meng, D.: Panflownet: A flow-based deep network for pan-sharpening. In: Proceedings of the IEEE/CVF International Conference on Computer Vision. pp. 16857--16867 (2023)

\bibitem{yang2017pannet}
Yang, J., Fu, X., Hu, Y., Huang, Y., Ding, X., Paisley, J.: Pannet: A deep network architecture for pan-sharpening. In: IEEE International Conference on Computer Vision. pp. 5449--5457 (2017)

\bibitem{yu2024deep}
Yu, H., Huang, J., Li, L., Zhao, F., et~al.: Deep fractional fourier transform. Advances in Neural Information Processing Systems  \textbf{36} (2024)

\bibitem{msdcnn}
Yuan, Q., Wei, Y., Meng, X., Shen, H., Zhang, L.: A multiscale and multidepth convolutional neural network for remote sensing imagery pan-sharpening. IEEE Journal of Selected Topics in Applied Earth Observations and Remote Sensing  \textbf{11}(3),  978--989 (2018)

\bibitem{Yuhas1992DiscriminationAS}
Yuhas, R.H., Goetz, A.F.H., Boardman, J.W.: Discrimination among semi-arid landscape endmembers using the spectral angle mapper (sam) algorithm (1992), \url{https://api.semanticscholar.org/CorpusID:126879175}

\bibitem{zamir2022restormer}
Zamir, S.W., Arora, A., Khan, S., Hayat, M., Khan, F.S., Yang, M.H.: Restormer: Efficient transformer for high-resolution image restoration. In: Proceedings of the IEEE/CVF conference on computer vision and pattern recognition. pp. 5728--5739 (2022)

\bibitem{zhou2022panformer}
Zhou, H., Liu, Q., Wang, Y.: Panformer: A transformer based model for pan-sharpening. In: 2022 IEEE International Conference on Multimedia and Expo (ICME). pp.~1--6. IEEE (2022)

\bibitem{zhou2022pan}
Zhou, M., Huang, J., Fang, Y., Fu, X., Liu, A.: Pan-sharpening with customized transformer and invertible neural network. In: Proceedings of the AAAI Conference on Artificial Intelligence. vol.~36, pp. 3553--3561 (2022)

\bibitem{zhou2022spatial}
Zhou, M., Huang, J., Yan, K., Yu, H., Fu, X., Liu, A., Wei, X., Zhao, F.: Spatial-frequency domain information integration for pan-sharpening. In: European Conference on Computer Vision. pp. 274--291. Springer (2022)

\bibitem{zhou2022mutual}
Zhou, M., Yan, K., Huang, J., Yang, Z., Fu, X., Zhao, F.: Mutual information-driven pan-sharpening. In: Proceedings of the IEEE/CVF Conference on Computer Vision and Pattern Recognition. pp. 1798--1808 (2022)

\bibitem{zhu2024vision}
Zhu, L., Liao, B., Zhang, Q., Wang, X., Liu, W., Wang, X.: Vision mamba: Efficient visual representation learning with bidirectional state space model. arXiv preprint arXiv:2401.09417  (2024)

\end{thebibliography}
\end{document}